%% file: main.tex
\documentclass[sigconf]{aamas} 

\usepackage{latexsym,amsmath}
\usepackage{hyperref}
\usepackage{caption}
\usepackage{multirow}
\usepackage{multicol}
\usepackage{algorithm}
\usepackage{algpseudocode}
\usepackage{makecell}
\usepackage{xcolor}

\newtheorem{definition}{Definition}

\usepackage{multirow} 
\usepackage{graphicx} 
\usepackage{balance} 

\setcopyright{ifaamas}
\acmConference[AAMAS '24]{Proc.\@ of the 23rd International Conference
on Autonomous Agents and Multiagent Systems (AAMAS 2024)}{May 6 -- 10, 2024}
{Auckland, New Zealand}{N.~Alechina, V.~Dignum, M.~Dastani, J.S.~Sichman (eds.)}
\copyrightyear{2024}
\acmYear{2024}
\acmDOI{}
\acmPrice{}
\acmISBN{}

\acmSubmissionID{477}

\newcommand\blfootnote[1]{%
  \begingroup
  \renewcommand\thefootnote{}\footnote{#1}%
  \addtocounter{footnote}{-1}%
  \endgroup
}

\title[OOD Detection in Reinforcement Learning ]{Rethinking Out-of-Distribution Detection for Reinforcement Learning: Advancing Methods for Evaluation and Detection}

\author{Linas Nasvytis}
\affiliation{
  \institution{Harvard University}
  \city{Cambridge, MA}
  \country{United States}}
\email{linasnasvytis@fas.harvard.edu}

\author{Kai Sandbrink}
\affiliation{
  \institution{University of Oxford}
  \city{Oxford}
  \country{United Kingdom}}
\email{kai.sandbrink@lmh.ox.ac.uk}

\author{Jakob Foerster}
\affiliation{
  \institution{University of Oxford}
  \city{Oxford}
  \country{United Kingdom}}
\email{jakob.foerster@engs.ox.ac.uk}

\author{Tim Franzmeyer$^\dagger$}
\affiliation{
  \institution{University of Oxford}
  \city{Oxford}
  \country{United Kingdom}}
\email{frtim@robots.ox.ac.uk}

\author{Christian Schroeder de Witt$^\dagger$}
\affiliation{
  \institution{University of Oxford}
  \city{Oxford}
  \country{United Kingdom}}
\email{cs@robots.ox.ac.uk}

\begin{abstract}
While reinforcement learning (RL) algorithms have been successfully applied across numerous sequential decision-making problems, their generalization to unforeseen testing environments remains a significant concern. 
In this paper, we study the problem of out-of-distribution (OOD) detection in RL, which focuses on identifying situations at test time that RL agents have not encountered in their training environments. 
We first propose a clarification of terminology for OOD detection in RL, which aligns it with the literature from other machine learning domains. 
We then present new benchmark scenarios for OOD detection, which introduce anomalies with temporal autocorrelation into different components of the agent-environment loop. 
We argue that such scenarios have been understudied in the current literature, despite their relevance to real-world situations. 
Confirming our theoretical predictions, our experimental results suggest that state-of-the-art OOD detectors are not able to identify such anomalies. 
To address this problem, we propose a novel method for OOD detection, which we call DEXTER (Detection via Extraction of Time Series Representations). 
By treating environment observations as time series data, DEXTER extracts salient time series features, and then leverages an ensemble of isolation forest algorithms to detect anomalies. 
We find that DEXTER can reliably identify anomalies across benchmark scenarios, exhibiting superior performance compared to both state-of-the-art OOD detectors and high-dimensional changepoint detectors adopted from statistics.
\end{abstract}


\newcommand{\dexterc}{DEXTER+C}

\newcommand{\BibTeX}{\rm B\kern-.05em{\sc i\kern-.025em b}\kern-.08em\TeX}

\makeatletter
\gdef\@copyrightpermission{
	\begin{minipage}{0.3\columnwidth}
		\href{https://creativecommons.org/licenses/by/4.0/}{\includegraphics[width=0.90\textwidth]{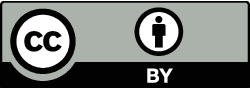}}
	\end{minipage}\hfill
	\begin{minipage}{0.7\columnwidth}
		\href{https://creativecommons.org/licenses/by/4.0/}{This work is licensed under a Creative Commons Attribution International 4.0 License.}
	\end{minipage}
	\vspace{5pt}
}
\makeatother

\begin{document}


\pagestyle{fancy}
\fancyhead{}


\maketitle 





\section{Introduction}

Deep reinforcement learning (RL) algorithms~\citep{DQN, PPO, SAC} have been applied to numerous sequential decision-making problems, ranging from robotics~\citep{todorov_mujoco_2012,andrychowicz_learning_2020} and nuclear fusion~\citep{degrave2022magnetic} to solar geoengineering~\citep{de_witt_stratospheric_2019}. 
However, their low reliability in situations that are not well represented in the training environment hinders the deployment of RL agents in safety-critical scenarios~\cite{chan2019measuring,packer2018assessing}. 
Such discrepant deployment situations are often referred to as out-of-distribution scenarios.
The task of identifying whether a given environment differs from the train-time environment is hence referred to as out-of-distribution (OOD) detection ~\cite{haider2023out}.
OOD detection constitutes an important desiderata for the deployment of RL agents in the real world, as reliable OOD detection would increase the safety of deployed AI agents by allowing contingency actions to be taken in unfamiliar or dangerous situations. 
Examples of such contingency options include the automatic parking of an autonomous car on the side of the road, or raising security escalations in cyber systems.\blfootnote{$^\dagger$ Equal supervision} 

The OOD detection framework assumes that the agent has access to data from the training process, which must be used to develop a mechanism for detecting OOD scenarios in the unknown deployment environment within a minimal amount of interactions. 
Such OOD deployment environments are often simulated by adding anomalies (e.g. sensor or process noise) to the train-time environments~\citep{mohammed2021benchmark,haider2023out,danesh2021out}.
In prior work, benchmark environments consider either injecting independent and identically distributed (i.i.d.) anomalies, such as Gaussian noise, or time-independent anomalies, such as changing gravity~\citep{mohammed2021benchmark,haider2023out,danesh2021out}. 
We argue that these scenarios fail to capture the \emph{temporally dependent} nature of many real-world anomalies.  
For example, in the case of an autonomous robot, any dust or smudge appearing on its camera lens would lead to a series of readings that contain systematic errors, instead of causing random misreadings of environmental data. 
Further, previous works~\citep{haider2023out,danesh2021out} employ decision rules based on per-step anomaly scores.
While such methods may prove empirically effective in some cases, we propose an information-theoretically grounded approach based on sequential hypothesis testing~\citep{wald_sequential_1945}, which results in a decision rule based on accumulated anomaly scores over multiple timesteps.

We start by introducing novel benchmark environments, which include temporally dependent anomalies, and find that state-of-the-art OOD detection methods fail in such scenarios.
To detect such anomalies, we introduce a novel detection mechanism called \textbf{D}etection via \textbf{Ex}traction of \textbf{T}ime S\textbf{e}ries \textbf{R}epresentations (DEXTER). This mechanism first extracts time-series features, which are then used with a random-forest algorithm to compute anomaly scores.
We leverage information-theoretically optimal sequential hypothesis testing techniques to derive a cumulative sum (CUSUM) detector using the full history of DEXTER's anomaly scores, which we refer to as \dexterc{}. 
Lastly, we evaluate DEXTER and \dexterc{} on a range of novel and standard benchmark environments and compare their performance to relevant baselines.
We find that DEXTER significantly outperforms state-of-the-art detectors across various metrics, including Area under the Receiver Operator Characteristic (AUROC) scores.
Importantly, DEXTER+C significantly decreases the number of timesteps needed to detect OOD scenarios.

Our work makes the following contributions:
\begin{itemize}
    \item We propose a clarification of the terminology of OOD detection in reinforcement learning. 
    \item We introduce new testing scenarios for OOD detection in reinforcement learning, which consider a broad class of disturbances focused on temporally-correlated noise. 
    \item We propose a new detector, which we refer to as DEXTER (\textbf{D}etection via \textbf{Ex}traction of \textbf{T}ime S\textbf{e}ries \textbf{R}epresentations), as well as a new decision rule, \dexterc{}, and show that these outperform state-of-the-art methods across relevant scenarios. 
\end{itemize}

\section{Related Work}
\label{sec:related_work}

\subsubsection{Algorithms for OOD Detection}
The research on out-of-distri-bution (OOD) detection in reinforcement learning~\citep{muller2022towards, haider2023out} is more limited compared to supervised and unsupervised learning. 
However, interest in the field has grown more recently. To our knowledge, \citet{sedlmeier2019uncertainty} outline the first practical method for OOD detection in reinforcement learning. The authors use epistemic uncertainty of the agent's actions to quantify the anomaly scores of different states, reasoning that epistemic uncertainty tends to be higher in areas of low data density. \citet{mohammed2021benchmark} introduce a benchmark to evaluate generalized OOD detection methods in reinforcement learning on three environments. In Cartpole and Pendulum, gravity is varied, while in Pong, noise is added to state observations. 
\citet{danesh2021out} propose a more extensive benchmark to test OOD detection, implementing different types of observational noise across seven classic RL environments. 
Additionally, they introduce a new OOD detector called the Recurrent Implicit Quantile Network (RIQN). 
At each time step, RIQN uses the current and prior states in the environment, $s_{1:t}$, to generate auto-regressive predictions for the next $\delta$ states, $s_{t+ 1: t+ \delta}$, then computes the difference between its predictions and the realized environment states, and uses this difference as the anomaly score for a given transition $(s_t, s_{t+1})$. 
The authors demonstrate that RIQN outperforms several baseline detectors across several of the proposed anomalous environments. 

The state-of-the-art OOD detection method, Probabilistic Ensemble Dynamics Model (PEDM), was proposed by \citet{haider2023out}, and consists of two components. 
First, a 1-step \textit{forward dynamics model} $f_{\theta}$ is learned, modeling the transition dynamics of the training environment, implemented as a Probabilistic Deep Neural Network Ensemble model~\citep{lakshminarayanan2017simple}.
For a given state and action pair $(s_t, a_t)$, this model predicts the next state in the environment as $s'_{t+1} = f_{\theta}(s_t, a_t)$.
Second, an \textit{anomaly score generator} is applied, which compares the predictions of the world dynamics model to outcomes in the test-time deployment environment, and generates an anomaly score for each transition. 

To measure performance in detecting anomalies applied to the observation space, \citet{haider2023out} use the environments from \citet{danesh2021out}. 
For the detection of anomalies applied to the transition dynamics of the RL environment, the authors propose four additional benchmark environments. They modify the classical Cartpole, HalfCheetah, Pusher, and Reacher environments by adding semantic anomalies, such as changing the gravity or multiplying the velocity applied to all body parts of the agent by a constant factor. The performance of PEDM is measured against the RIQN algorithm \citep{danesh2021out}, as well as several other benchmark models. For each algorithm, the performance is measured according to AUROC. Based on the results, PEDM outperforms the other detection algorithms across almost all newly-proposed environments with anomalous transition dynamics. Moreover, within the environments from \citet{danesh2021out}, PEDM is tied with the RIQN model for the best performance (with RIQN exhibiting significantly worse performance on the new benchmarks).

\subsubsection{Limitations of current OOD detection approaches.}
\label{sec:limitations}


We argue that current approaches to OOD detection suffer from three weaknesses.
First, most of the literature of OOD detection in RL makes the simplifying assumption of injecting independent and identically distributed (i.i.d.) noise across different timesteps in the environment \citep{danesh2021out,haider2023out,mohammed2021benchmark}, which can hence be reliably detected with one-step detection approaches. However, relevant real-world scenarios will likely feature more complex disturbances, which have temporally correlated anomalies.
Second, these approaches rely on anomaly detection via prediction error, computing anomaly scores based on the prediction error from a forward-dynamics model trained on training samples. 
Such prediction models are unlikely to detect temporally-correlated anomalies which may only have a minor effect on the transition dynamics.
Third, the decision rules for online out-of-distribution detection rely only on the individual anomaly scores observed at each time step, instead of taking into account the full history of anomaly scores. 
While such approaches may prove empirically effective, they neglect information that would be used in information-theoretically grounded sequential hypothesis testing~\citep{wald_sequential_1945}.

\begin{figure}[t]
    \centering
    \includegraphics[width=1.0\linewidth]{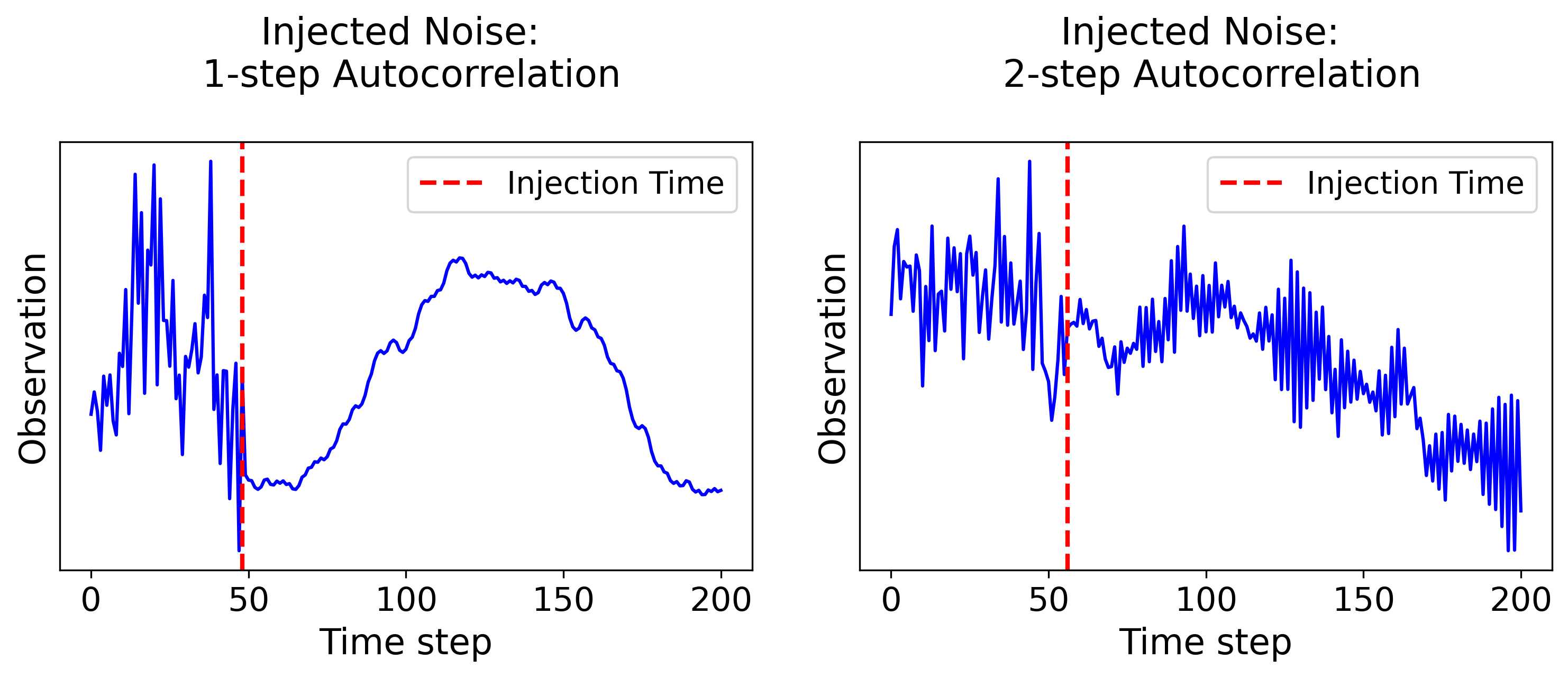}
    \caption[Illustration of temporally autocorrelated anomalies]{Illustration of temporally autocorrelated anomalies. Left: at injection time (t = 48), noise applied to the observation changes from no correlation to 1-step autocorrelation. Right: at injection time (t = 56), noise changes from no correlation to 2-step autocorrelation.}
    \label{fig:acorr}
\end{figure}

\subsubsection{Sequential Hypothesis Testing.}
Beyond machine learning, the task of online detection of whether a sample of time series data differs from a predefined distribution has also been analyzed in statistics, primarily in the field of changepoint detection (CPD).
Sequential hypothesis testing, as pioneered by Wald in his sequential probability ratio test (SPRT) \citep{wald_sequential_1945}, aims to discern between two hypotheses in an online manner, using as few samples as possible. 
CUSUM methods, a popular technique in CPD, are designed to detect shifts in the mean or variance of a process \citep{page_continuous_1954}. 
CUSUM methods focus on capturing cumulative information over time, providing a mechanism to identify time-dependent anomalies. 

\section{Background and Notation}
\label{sec:background}
A Markov Decision Process (MDP) is defined by a tuple $(S, A, P, R)$. Here, $S$ is a finite set of states, $A$ is a finite set of actions, $P: S \times A \times S \rightarrow [0,1]$ defines the state transition probabilities, and $R: S \times A \rightarrow \mathbb{R}$ is the reward function. The goal of the agent at each time step is to maximize the cumulative sum of discounted future rewards $\sum_{t}^{t \rightarrow \infty} \gamma^t R(s_t, a_t)$, where $\gamma \in [0,1]$ is the discount rate.

In the general machine learning literature -- which includes supervised learning, unsupervised learning, and reinforcement learning -- there are several terms used to describe the task of detecting if a new sample of data differs from a specified training set. 
The terms can include out-of-distribution (OOD) detection, anomaly detection, novelty detection, and outlier detection. 
To clarify the terminology, \citet{yang2021generalized} provide an in-depth literature review of this topic, focusing primarily on supervised learning, where each data point contains an input-label pair ($x$,$y$) $\in X \times Y$, 
where \(X\) is the input (sensory) space, and \(Y\) the label (semantic) space. 
A data distribution is defined as a joint distribution \(P(X, Y)\) over \(X \times Y\). 
A distribution shift can occur in either the marginal distribution \(P(X)\), or both \(P(Y)\) and \(P(X)\). 
As noted by \citet{yang2021generalized}, a shift in \(P(Y)\) naturally triggers shift in \(P(X)\). 
Denote the distributions of normal and anomalous data points by $P$ and $P'$, respectively. 
\textbf{Covariate shift (sensory) anomalies} occur when the inputs of normal data points are drawn from an in-distribution \(P(X)\), whereas inputs of anomalies are drawn from out-of-distribution \(P'(X)\), such that \(P(X) \neq P'(X)\). 
However, no distribution shift occurs in the label space: \(P(Y) = P'(Y)\). 
\textbf{Semantic shift anomalies} occur when a distributional shift occurs in the label space, such that \(P(Y) \neq P'(Y)\). Following this distinction, the authors define: 
\begin{enumerate}
    \item \textbf{Sensory anomaly detection} as the task of detecting covariate shift anomalies, i.e. samples from \(P'(X)\).
    \item \textbf{Semantic anomaly detection} as the task of detecting semantic shift anomalies, i.e. samples from \(P'(Y)\).
    \item \textbf{Out-of-distribution (OOD) detection} as the sub-domain within semantic anomaly detection, where in-distribution samples are drawn from multiple classes (i.e. label space $Y$ is not binary).
    \item \textbf{Generalized out-of-distribution detection} as the general task of detecting all anomalies, i.e. both sensory and semantic anomalies.
\end{enumerate}

The main issue is that the training and testing data in reinforcement learning does not contain a label space $Y$. 
As a result, the terms \textit{out-of-distribution detection} and \textit{anomaly detection} are often used interchangeably \citep{haider2023out}, and some of their definitions seem to conflict with each other. A detailed overview of such terminological discussions on OOD detection in reinforcement learning can be found in the Appendix.

\section{Terminology of OOD Detection in Reinforcement Learning}
\label{sec:terminology}
Given terminological discussions on how to label different types of anomalies for OOD detection in reinforcement learning, we propose terminology adapted from the framework introduced by \citet{yang2021generalized}, while incorporating insights from existing literature from \citet{danesh2021out} and \citet{haider2023out}. 
We first differentiate between two kinds of anomalies based on their effects on the MDP:
First, \textbf{sensory anomalies} change the observation that the agent receives (e.g. adding observational noise), while leaving the underlying environment dynamics unchanged. Hence, these are closely related to covariate shift anomalies defined in \citet{yang2021generalized}. In contrast,
\textbf{semantic anomalies} change the underlying environment dynamics (e.g. changing the gravity in the environment). Hence, these are very similar to semantic shift anomalies defined in \citet{yang2021generalized}.
We further define \textbf{Generalized Out-of-Distribution Detection} as the task of detecting either of the two.

\begin{definition}
\textbf{Sensory anomaly detection} in reinforcement learning refers to the task of identifying sensory anomalies. A sensory anomaly is a perturbation to the reinforcement learning environment, which changes the observations $O$ that the agent receives, but leaves the underlying environment dynamics unchanged. If the change in observation leads the environment to become partially observable, such an anomaly changes the underlying Markov Decision Process (MDP) to a Partially Observable Markov Decision Process (POMDP).
\end{definition}

\begin{definition}
\textbf{Semantic anomaly detection} in reinforcement learning refers to the task of identifying semantic anomalies. Semantic anomalies are perturbations to the reinforcement learning environment, which change the transition function of a Markov Decision Process $P(S'| S, A)$ by changing the environment dynamics.
\end{definition}

\begin{definition}
\textbf{Generalized out-of-distribution detection} in reinforcement learning refers to the task of identifying any type of anomaly in the environment, hence including sensory anomaly detection and semantic anomaly detection.
\end{definition}

The proposed terminology unites the task of detecting all types of anomalies in reinforcement learning environments under the single term of generalized out-of-distribution detection, while drawing a meaningful distinction between the two major types of anomalies that could exist in an environment. 
This is especially relevant as some recent works have been inaccurate about the changes to the MDP that the anomalies introduce (see Appendix for a discussion).

\section{Novel Testing Scenarios for OOD Detection}
\label{sec:scenarios}
We now propose new benchmark environments for generalized OOD detection that contain both sensory and semantic anomalies.
In contrast to the benchmark environments used in previous works, these environments contain temporally correlated anomalies, which is achieved by generating and adding noise using the autoregressive (AR) model \citep{box2015time}. 

\subsubsection{Autocorrelated noise patterns}
To create custom benchmark environments with different types of noise correlations, we use an AR model of order $p$, denoted by AR(p), where the noise value $Y_t$ of a series at any point in time is linearly dependent on its own $p$ past values: 
$$
Y_t = \mu + \phi_1 Y_{t-1} + \phi_2 Y_{t-2} + \cdots + \phi_p Y_{t-p} + \varepsilon_t
$$

Using this AR model, we implement noise with three different types of autocorrelations: 
\begin{enumerate}
    \item \textbf{No correlation}, in which case the noise is not correlated across time, i.e., containing white noise, given as  $Y_t = \epsilon_t$.
    \item \textbf{1-step correlation}, in which case the noise is autocorrelated across each step, hence given as $Y_t = \phi_1 Y_{t-1} + \epsilon_t$.
    \item \textbf{2-step correlation}, where the noise is autocorrelated only at every second step (i.e., step 2, step 4), in which case, $Y_t = \phi_2 Y_{t-2} + \epsilon_t$.
\end{enumerate}

\begin{figure}[t]
    \centering
    \includegraphics[width=1.0\linewidth]{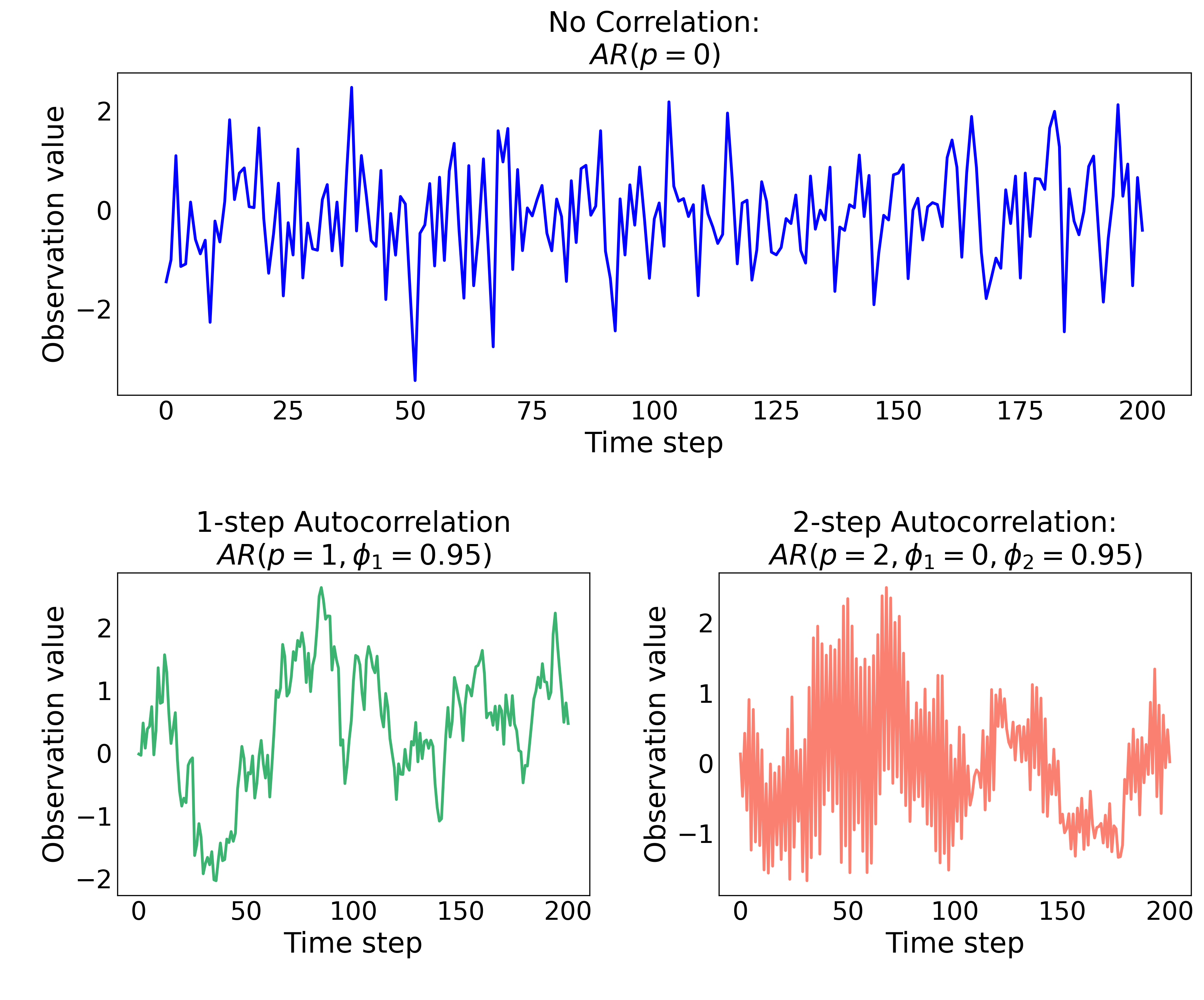}
    \caption[Illustrations of Autoregressive (AR) model with different parameters]{Illustrations of Autoregressive (AR) model with parameters for no correlation (top), 1-step autocorrelation (bottom-left), and 2-step autocorrelation (bottom-right), which is used to create three different types of noise in new testing scenarios.}
    \label{fig:AR}
\end{figure}

Examples of the three types of noise for correlation coefficients are displayed for $\phi=0.95$ in Figure \ref{fig:AR}.

Based on the autocorrelated noise patterns introduced above, we introduce three new benchmark environments.

\subsubsection{ARTS: Autoregressive Time Series environments}
Autoregressive Time Series (ARTS) serves as our baseline environment. At the beginning of each episode, we generate a 1-dimensional vector using the AR process, with the number of elements equal to the maximum number of episode steps. At each time step $t$, the agent receives a \textit{1-dimensional observation} $o_t$ that contains the noise generated by the AR process, while the environment state is treated as constant.

\subsubsection{ARNO: Autoregressive Noised Observation environments}
In the Autoregressive Noised Observation (ARNO) setting, we introduce \textbf{sensory anomalies}. At the beginning of the episode, we generate a noise matrix, where each row of the matrix is a time series, independently drawn from the AR Process. The number of rows in the matrix corresponds to the dimension of a single observation in the environment.

For each step $t$, the agent takes an action $a_t$, which is passed to the environment's state transition function to generate the next state: $s_{t+1} \sim P(s_{t+1} | s_t, a_t)$. \textit{After} each state transition, the noise vector sliced from the matrix is added to the state $s_{t+1}$, transforming it into an observation $o_{t+1}$. This simulates a scenario where the environment's underlying state undergoes a distortion before being presented to the agent, similar to the effects of a camera glitch or a sensory failure.

\subsubsection{ARNS: Autoregressive Noised State environments}
In Autoregressive Noised State (ARNS) environments, we introduce \textbf{semantic anomalies}. Similarly to ARNO, we generate the AR noise matrix at the beginning of each episode, with the same shape. At each step $t$, the agent takes an action $a_t$, which is passed to the environment's state transition function. However, \textit{before} the transition, a noise vector sliced from the matrix is applied to each dimension of the state vector, effectively changing the transition function. This simulates a scenario where, for example, the underlying physics or rules of the environment are unpredictably changed due to systematic issues.

\section{DEXTER: Detection via Extraction of Time Series Representations}\label{sec:DEXTER}
We now move on to discuss our proposed algorithm for out-of-distribution detection in reinforcement learning, which we refer to as \textbf{DEXTER} (Detection via Extraction of Time-series Representations). The model is composed of two components: 
first, a \textbf{feature extractor} $f(s_1, ... s_n)$ that extracts relevant time series features from a given time series data; and
second, an \textbf{anomaly detector} $h(f(s_1, ... s_n)$, which we implement as an ensemble of isolation forest models that takes as an input the extracted features, and outputs an anomaly score for the time series.

\subsubsection{Feature extractor $f$}
Given a sample of states $s_0, s_1, \ldots s_t$, DEXTER first extracts the relevant time series features from each state dimension. 
The goal behind the proposed detector is to be anomaly agnostic: Instead of choosing features that could detect a specific anomaly (e.g. autocorrelations), we aim to extract a diversity of features to maximize the number of anomalies DEXTER could detect. 
For this purpose, we use the ~\texttt{tsfresh} feature extractor~\citep{christ2018time}, which captures features of a diverse set of statistics of the time series, including fundamental descriptive statistics (e.g. number of peaks, minimum, maximum, and median values), autocorrelation statistics (e.g. autocorrelation coefficients for $k$ lags; partial autocorrelation function \citep{box2015gm} at the given lag $k$), advanced features (e.g. descriptive statistics of the absolute Fourier transform spectrum~\citep{marple1989digital} and approximate entropy~\citep{delgado2019approximate}). 

\subsubsection{Isolation Forest Algorithm $h$}
Next, we fit an ensemble of isolation forest algorithms, which predicts the probability of an anomaly based on the extracted features.
The isolation forest algorithm is an ensemble-based unsupervised machine learning method designed for anomaly detection, introduced by~\citet{liu2008isolation}. Given a training set, the algorithm first constructs multiple random trees to partition data points. For each tree, the algorithm selects a sample of training data, and recursively partitions the sample of data points by randomly selecting a split value that falls between the minimum and maximum values of a given attribute (i.e. dimension). Once an ensemble of such isolation trees is constructed, a test data point is passed down each of these trees. The path length it takes to isolate the data point is averaged over the trees to produce an anomaly score for the data point. The algorithm relies on the assumption that anomalous points will be more easily separable from the rest of the sample \citep{liu2012isolation}.

Isolation forest has the following beneficial properties: First, it is an \textbf{unsupervised} method, which is essential as we are not provided anomalous data and hence must train an anomaly-agnostic detector. 
Second, it has \textbf{linear time complexity} \citep{liu2008isolation}, allowing to scale to high-dimensional data. 
Third, it allows for \textbf{anomaly scoring}, enabling a nuanced interpretation of how "anomalous" a data point is, instead of providing a mere binary classification. 

\subsubsection{Training and testing.}
An overview of DEXTER Anomaly Score computation is displayed in Algorithm~\ref{algo:DEXTER}.
We will denote the training data as a sample $D = \{s_n, a_n, s_{n+1}\}^{N}_{n=1}$. 

\paragraph{Training.} First, DEXTER partitions $D$ into windows of size \( W \) states. 
For each window $w_i$, the model extracts time-series features along each dimension $d$, $1 \leq d \leq m$, to obtain $f_{i, d}$. 
Once feature extraction for all windows is complete: for each dimension $d$, all extracted features are concatenated to make $f_d$. 
This aggregate feature set is used to fit the ensemble of isolation forest models, where each model is assigned to a different state dimension, resulting in $m$ total models. 
For each model in the ensemble $IF_d$, the aim is to instill an understanding of the predominant patterns present in the given dimension of in-distribution observations.

\paragraph{Dexter Anomaly Score Computation.} 
For a given timestep T, DEXTER collates the last \( W \) states to constitute a window, and extracts features \( f_d' \) along each dimension in $m$. 
The ensemble of trained isolation forest models is then utilized to compute anomaly scores for these features. 
The concluding anomaly score for a timestep is calculated as the arithmetic mean of the scores spanning all the dimensions in \(m\).

\subsection{Sequential Hypothesis Testing with \dexterc{}}
The previously described DEXTER algorithm outputs an anomaly score after each transition in the test-time deployment environment.
We now introduce \dexterc{} (short for DEXTER+CUSUM), which uses the information-theoretic hypothesis testing CUSUM algorithm~\citep{page_continuous_1954} to derive a decision rule for when to classify a test-time deployment as OOD.
\dexterc{} is detailed in Algorithm~\ref{algo:DEXTERC}.
First, the average DEXTER anomaly score $\bar{A}$ for a held-out set of training transitions is computed.
Then, the CUSUM detection threshold $\tau$ is computed by evaluating CUSUM on held-out training transitions and choosing $\tau$ such that a targeted False Positive Rate ($FPR$) is achieved on the set of held-out transitions.
This tunable targeted $FPR$ indicates the ratio of anomaly-free episodes that are falsely classified as OOD.
Note that both steps only require access to training transitions and that the only hyperparameter is the targeted $FPR$.
At test time, an online CUSUM detector with detection threshold $\tau$ is employed.
This detector updates the CUSUM score $S_t$ based on the previous timestep's score and the current DEXTER anomaly score $A_t$.
If the CUSUM score $S_t$ exceeds $\tau$, an OOD scenario is detected and the execution is halted.

\begin{algorithm}[t]
\caption{DEXTER}
\label{algo:DEXTER}
\begin{algorithmic}
\Require State dimensions \( m \), window size \( W \), policy \( \pi \), dataset \( D \) of training transitions

\Statex
\Statex \underline{Training}
\State Initialize ensemble of Isolation Forests \(\text{IF} = \{\text{IF}_1, ..., \text{IF}_m\} \)
\State Partition \( D \) into windows \({w_1, ..., w_N}\) of size \( W \)

\For{each \( w_i \) in windows}
    \For{dimension \( d \) from 1 to \( m \)}
        \State Extract time series features \( f_{i,d} \)
    \EndFor
\EndFor

\For{dimension \( d \) from 1 to \( m \)}
    \State Form \( f_d \) by concatenating \( f_{i,d} \)
    \State Train \( IF_d \) using \( f_d \)
\EndFor

\Statex
\Statex \underline{\textbf{DEXTER Anomaly Score Computation}}
\For{Time \( t \) from 0 to \( T \)}
    \State Action \( a_t \gets \pi(s_t) \), observe \( s_{t+1} \)
    \State Update window with \( s_{t+1} \)
    \For{dimension \( d \) from 1 to \( m \)}
        \State Extract features \( f_d \)
        \State Compute score \( a_d \) with \( \text{IF}_d \)
    \EndFor
    \State Set \( A_T \) as average over all \( a_d \)
\EndFor
\end{algorithmic}
\end{algorithm}

\begin{algorithm}[t]
\caption{\dexterc{}}
\label{algo:DEXTERC}
\begin{algorithmic}
\Require targeted $FPR$, state dimensions \( m \), window size \( W \), policy \( \pi \), validation dataset \( D \) of training transitions

\Statex 
\Statex \underline{Compute Mean Anomaly Score $\bar{A}$ using Validation Set}

\State Compute DEXTER anomaly scores for \( \frac{1}{2} D \), storing in \{ $A_t$ \}
\State \( \bar{A} \gets \) mean of \{ $A_t$ \}


\Statex 
\Statex \underline{Compute CUSUM Threshold $\tau$ using Validation Set}

\State \( S_{max\_list} \gets [] \)

\For{each episode \( ep \) in \( \frac{1}{2} D \)}
    \State \( S_0 \gets 0, S_{max} \gets 0 \)
    \For{each score \( A_t \) in \( ep \)}
        \State \( S \gets S + A_t - \bar{A} \)
        \State \( S_{max} \gets \max(S, S_{max}) \)
    \EndFor
    \State Append \( S_{max} \) to \( S_{max\_list} \)
\EndFor
\State \( \tau \gets \) 1-\( FPR \) percentile of \( S_{max\_list} \)

\Statex 
\Statex \underline{\textbf{\dexterc{} Out-of-Distribution Detection}}

\State \( S_0 \gets 0 \)
\For{Time \( t \) from 0 to \( T \)}
    \State Compute \( A_t \) using DEXTER Anomaly Score Computation
    \State \( S_t \gets \max(0, S_{t-1} + A_t - \bar{A}) \)
    \If{\( S_t \) > \( \tau \)}
        \State Raise out-of-distribution alert
        \State Break
    \EndIf
\EndFor
\end{algorithmic}
\end{algorithm}

\section{Empirical Evaluation}
\label{sec:experiment}
We first focus on the newly introduced benchmark scenarios ARNO, ARNS, and ARTS, which contain temporally correlated anomalies. We implement ARNO scenarios with three levels of noise (described in Section \ref{sec:eval_new_scenarios}) in three different environments -- Cartpole, Reacher, and Acrobot~\citep{gym}. We implement ARNS scenarios with analogous levels of noise on Cartpole and Reacher, as the implementation on Acrobot leads to inconsistent agent policies, which are described in more detail in the Appendix. Lastly, as the ARTS environment does not contain a reward signal, we only implement a single noise level. The codebase and supplementary materials are publicly available at: \url{https://github.com/LinasNas/DEXTER}.

We afterward consider the common benchmark scenarios introduced by \citet{haider2023out}, which contain either i.i.d anomalies or time-independent anomalies (five scenarios in total), two noise levels for each anomaly, and two different environments (Cartpole and Reacher), resulting in 20 different evaluation frameworks.

Previous works~\citep{haider2023out,danesh2021out} focus on AUROC computed for per-transition anomaly scores as the main metric for detector performance.
However, AUROC scores do not yield a decision rule for classifying observation as OOD.
We hence also consider the required timesteps to classify the anomaly-containing test-time environment as OOD as a metric for detector performance, we refer to this metric as \emph{Detection Time}.

\subsection{Evaluations for ARTS, ARNS and ARNO Scenarios}
\label{sec:eval_new_scenarios}
\paragraph{Noise levels in each environment}
In ARNO and ARNS environments, we apply three different levels of noise to generate the anomalies. 
The levels are classified as Light, Medium, and Strong, based on their effect on the agent's reward in the environment. 
The procedure to choose these levels is as follows. 
First, we normalize the noise magnitude by the standard deviation of that dimension's observation.
Then, we apply \textit{uncorrelated noise} of different magnitudes to the environment and train a reinforcement learning agent using Proximal Policy Optimization (PPO) \citep{schulman2017proximal} in discrete action-space environments, and Twin Delayed Deep Deterministic (TD3)\cite{fujimoto2018addressing} algorithm for continuous action-space environments, until it converges to a stable cumulative episodic reward. 
We then measure how much, averaged over 50 episodes, the reward differs from the reward achieved when an agent's policy is optimized in an undisturbed environment. 
Lastly, in each environment, we identify three magnitudes of noise, which we classify as:

\begin{itemize}
    \item Light noise: reduces $\sum_{t=0}^{T} r_t$ by $\sim 1\%$
    \item Medium noise: reduces $\sum_{t=0}^{T} r_t$ by $\sim 25\%$
    \item Strong noise: reduces  $\sum_{t=0}^{T} r_t$ by $\sim 50\%$
\end{itemize}

\subsubsection{Evaluation Procedure}
Our experimental evaluation follows recent frameworks~\citep{haider2023out,danesh2021out}. 
We first apply uncorrelated noise of each of the three levels. 
Then for each noise level, we train a policy $\pi_A(a|s)$ until it is optimized for the episodic task with episode length $H \in \mathbb{N}$. 

In each episode, we introduce an anomaly at a random time $t_a \in (t_0 + 5, t_H - 5)$, and apply this anomaly until the end of the episode. 
That anomaly changes the correlation structure of noise from \textit{no correlation} to either \textit{1-step} or \textit{2-step} correlation, as it can be observed in Figure 1.
We test each of the two cases separately. 
All transitions before the anomaly is applied, $[(s_{t_0}, s_{t_1}), \ldots, (s_{t_a-1}, s_{t_a})]$ are labeled as in-distribution, and all transitions after, $[(s_{t_a}, s_{t_{a+1}})$ $\ldots, (s_{t_{H-1}}, s_{t_H})]$ are labelled as anomalous. 

To account for the effects of initialization and varying injection times, we repeat this procedure for multiple random point time points and different initial environment states. 
Therefore, in expectation, we obtain a balanced dataset. 

Importantly, the noise level throughout the episode is constant, since  after the injection time, we simply change the correlation structure of the noise, but not its magnitude. 
The experiments are implemented in such a way since we are specifically interested in whether the detectors can identify structural changes to the noise, rather than detect the sudden emergence of noise itself, unlike most prior benchmark environments~\citep{haider2023out,danesh2021out}. 

\begin{table}[t]
    \caption{ARTS scenarios: Detector performance (AUC above, Detection time below).}
    \label{tab:ARTS}
    \centering
    \fontsize{8pt}{11pt}\selectfont
    \resizebox{0.7\linewidth}{!}{\begin{tabular}{lccc}
        \hline
        & & 1-step & 2-step \\
        \hline
        \multirow{4}{*}{AUROC $\uparrow$} 
        & CPD: OCD & 0.79 & 0.77  \\
        & CPD: Chan & 0.64 & 0.64 \\
        & PEDM & 0.51 & 0.5   \\
        & DEXTER & \textbf{0.89} & \textbf{0.83}  \\
        \hline
        \hline
        \multirow{2}{*}{Det. Time $\downarrow$} & PEDM+C & 200.0 & 200.0  \\
        & \dexterc{}                                      & \textbf{19.8} & \textbf{28.0}   \\
        \hline
    \end{tabular}}
\end{table}


\begin{table*}[t]
    \caption{ARNO scenarios: Detector performance (AUC above, Detection time below).}
    \label{tab:ARNS}
    \centering
    \fontsize{8pt}{11pt}\selectfont
    \resizebox{\linewidth}{!}{
    \begin{tabular}{l|c|cc|cc|cc|cc|cc|cc|cc|cc|cc}
        \hline
        & & \multicolumn{6}{c|}{Cartpole} & \multicolumn{6}{c|}{Acrobot} & \multicolumn{6}{c}{Reacher} \\
        \cline{3-20}
        & & \multicolumn{2}{c|}{Light Noise} & \multicolumn{2}{c|}{Medium Noise} & \multicolumn{2}{c|}{Strong Noise} & \multicolumn{2}{c|}{Light Noise} & \multicolumn{2}{c|}{Medium Noise} & \multicolumn{2}{c|}{Strong Noise} & \multicolumn{2}{c|}{Light Noise} & \multicolumn{2}{c|}{Medium Noise} & \multicolumn{2}{c}{Strong Noise} \\
        \hline
        & Detector & 1-step & 2-step & 1-step & 2-step & 1-step & 2-step & 1-step & 2-step & 1-step & 2-step & 1-step & 2-step & 1-step & 2-step & 1-step & 2-step & 1-step & 2-step \\
        \hline
        \multirow{4}{*}{AUROC $\uparrow$} 
        & CPD: OCD & 0.67 & 0.69 & 0.76 & 0.72 & 0.78 & 0.73 & 0.64 & 0.65 & 0.75 & 0.76 & 0.8 & 0.79 & 0.51 & 0.51 & 0.51 & 0.51 & 0.52 & 0.52 \\
        & CPD: Chan & 0.69 & 0.68 & 0.72 & 0.75 & 0.75 & 0.73 & 0.62 & 0.59 & 0.76 & 0.71 & 0.86 & 0.77 & 0.51 & 0.51 & 0.52 & 0.52 & 0.53 & 0.53 \\
        & PEDM & 0.55 & 0.62 & 0.6 & 0.51 & 0.6 & 0.55 & 0.57 & 0.54 & 0.52 & 0.54 & 0.5 & 0.53 & \textbf{0.81} & 0.5 & 0.84 & 0.51 & 0.87 & 0.5 \\
        & DEXTER & \textbf{0.81} & \textbf{0.85} & \textbf{0.89} & \textbf{0.9} & \textbf{0.93} & \textbf{0.9} & \textbf{0.74} & \textbf{0.71} & \textbf{0.96} & \textbf{0.91} & \textbf{0.99} & \textbf{0.95} & 0.67 & \textbf{0.6} & \textbf{0.91} & \textbf{0.63} & \textbf{0.97} & \textbf{0.61} \\
        \hline
        \hline
        \multirow{2}{*}{Det. Time $\downarrow$} 
        & PEDM+C & 138.8 & 143.2 & 133.0 & 79.3 & 199.1 & 183.1 & 163.9 & 199.2 & 79.7 & 197.2 & 177.2 & 169.9 & \textbf{21.3} & 200.7 & \textbf{19.3} & \textbf{193.6} & 18.1 & 200.8 \\
        & DEXTER+C & \textbf{23.2} & \textbf{15.2} & \textbf{13.1} & \textbf{19.9} & \textbf{14.85} & \textbf{14.7} & \textbf{32.1} & \textbf{42.5} & \textbf{12.7} & \textbf{13.9} & \textbf{8.0} & \textbf{11.5} & 60.9 & \textbf{72.1} & 20.0 & 197.1 & \textbf{12.4} & \textbf{110.6} \\
        \hline
    \end{tabular}}
\end{table*}

\begin{table*}[t]
    \caption{ARNS scenarios: Detector performance (AUC above, Detection time below).}
    \label{tab:ARNO}
    \centering
    \fontsize{8pt}{11pt}\selectfont
    \resizebox{0.75\linewidth}{!}{
    \begin{tabular}{l|c|cc|cc|cc|cc|cc|cc}
        \hline
        & & \multicolumn{6}{c|}{Cartpole} & \multicolumn{6}{c}{Reacher} \\
        \cline{3-14}
        & & \multicolumn{2}{c|}{Light Noise} & \multicolumn{2}{c|}{Medium Noise} & \multicolumn{2}{c|}{Strong Noise} & \multicolumn{2}{c|}{Light Noise} & \multicolumn{2}{c|}{Medium Noise} & \multicolumn{2}{c}{Strong Noise} \\
        \hline
        & Detector & 1-step & 2-step & 1-step & 2-step & 1-step & 2-step & 1-step & 2-step & 1-step & 2-step & 1-step & 2-step \\
        \hline
        \multirow{4}{*}{AUROC $\uparrow$} 
        & CPD: OCD & 0.66 & 0.66 & 0.68 & 0.68 & 0.67 & 0.68 & 0.51 & 0.51 & 0.51 & 0.51 & 0.51 & 0.51 \\
        & CPD: Chan & 0.67 & 0.68 & 0.68 & 0.69 & 0.68 & 0.7 & 0.51 & 0.51 & 0.51 & 0.51 & 0.51 & 0.51 \\
        & PEDM & 0.66 & 0.64 & 0.63 & 0.61 & 0.59 & 0.56 & 0.52 & 0.51 & \textbf{0.55} & 0.55 & 0.51 & 0.5 \\
        & DEXTER & \textbf{0.73} & \textbf{0.73} & \textbf{0.88} & \textbf{0.8} & \textbf{0.84} & \textbf{0.77} & \textbf{0.56} & \textbf{0.62} & 0.51 & \textbf{0.7} & \textbf{0.55} & \textbf{0.67} \\
        \hline
        \hline
        \multirow{2}{*}{Det. Time $\downarrow$} 
        & PEDM+C & 33.0 & 44.5 & 53.1 & 50 & 44.5 & 125.3 & \textbf{189.4} & 197.7 & 201.0 & 201.0 & \textbf{195.8} & 197.2 \\
        & DEXTER+C & \textbf{32.2} & \textbf{16.9} & \textbf{17.8} & \textbf{37.8} & \textbf{24.0} & \textbf{27.4} & 199.0 & \textbf{95.1} & \textbf{193.6} & \textbf{61.1} & 198.7 & \textbf{66.7} \\
        \hline
    \end{tabular}}
\end{table*}

\subsubsection{DEXTER and \dexterc{}}
We implement DEXTER (described in Algorithm~\ref{algo:DEXTER}) with a window size of $W = 10$ timesteps, since it strikes a balance between a sample that is long enough to allow for meaningful time series feature extraction, while being short enough to allow for quick anomaly detection. 
We implement \dexterc{} (described in Algorithm~\ref{algo:DEXTERC}) with a target $FPR$ of 1\%.


\begin{table*}[t]
    \caption{Benchmark scenarios: Detector performance (AUC above, Detection time below).}
    \label{tab:bench}
    \centering
    \fontsize{10pt}{14pt}\selectfont
    \resizebox{\linewidth}{!}{
    \begin{tabular}{l|c|cc|cc|cc|cc|cc|cc|cc|cc|cc|cc}
        \hline
        & & \multicolumn{10}{c|}{Cartpole} & \multicolumn{10}{c}{Reacher} \\
        \cline{3-22}
        & & \multicolumn{2}{c|}{Action Fact.} & \multicolumn{2}{c|}{Action Noise} & \multicolumn{2}{c|}{Action Offset} & \multicolumn{2}{c|}{Body M. Fact.} & \multicolumn{2}{c|}{Force Vector} & \multicolumn{2}{c|}{Action Fact.} & \multicolumn{2}{c|}{Action Noise} & \multicolumn{2}{c|}{Action Offset} & \multicolumn{2}{c|}{Body M. Fact.} & \multicolumn{2}{c}{Force Vector} \\
        \hline
        & Detector & Minor & Severe & Minor & Severe & Minor & Severe & Minor & Severe & Minor & Severe & Minor & Severe & Minor & Severe & Minor & Severe & Minor & Severe & Minor & Severe \\
        \hline
        \multirow{2}{*}{AUROC $\uparrow$} 
        & PEDM         & 0.59 & \textbf{0.94} & 0.66 & \textbf{0.98} & \textbf{0.82} & \textbf{1.0} & 0.6 & \textbf{0.9} & 0.59 & \textbf{1.0} & 0.62 & \textbf{0.95} & 0.59 & \textbf{0.99} & 0.61 & \textbf{0.98} & 0.64 & 0.56 & 0.74 & \textbf{0.98} \\
        & DEXTER       & \textbf{0.76} & 0.64 & \textbf{0.75} & 0.55 & 0.76 & 0.71 & \textbf{0.75} & 0.66 & \textbf{0.72} & 0.56 & \textbf{0.73} & 0.69 & \textbf{0.72} & 0.62 & \textbf{0.74} & 0.66 & \textbf{0.71} & \textbf{0.73} & \textbf{0.76} & 0.69 \\
        \hline
        \hline
        \multirow{2}{*}{Det. Time $\downarrow$} 
        & PEDM+C       & 153.1 & \textbf{1.7} & 76.3 & \textbf{1.7} & 200.0 & \textbf{1.1} & 151.3 & \textbf{2.9} & 127.1 & \textbf{1.1} & 199.6 & 196.6 & 200.8 & \textbf{52.7} & 200.5 & 192.3 & 199.3 & 201.0 & 200.1 & 200.8 \\
        & DEXTER+C     & \textbf{44.8} & 50.4 & \textbf{42.1} & 69.4 & \textbf{44.0} & 52.8 & \textbf{46.2} & 50.9 & \textbf{48.2} & 200.0 & \textbf{53.6} & \textbf{53.3} & \textbf{51.9} & 129.9 & \textbf{46.8} & \textbf{65.6} & \textbf{53.7} & \textbf{52.1} & \textbf{44.0} & \textbf{55.2} \\
        \hline
    \end{tabular}}
\end{table*}

\subsubsection{Probabilistic Ensemble Dynamics Model (PEDM)} We implement the PEDM from \citet{haider2023out} as the state-of-the-art benchmark for generalized out-of-distribution detection. For a fair comparison, we further also implement a CUSUM-based PEDM detector analogously to \dexterc{}, which we refer to as PEDM-C.

\subsubsection{Changepoint detectors} In the newly introduced scenarios, we also implement change point detectors from \citet{chen2022high}, which the authors introduce under the name OCD, as well as another changepoint detector proposed by \citet{chan2017optimal}. 
OCD conducts likelihood ratio tests against simple alternatives of different scales in each coordinate, and then aggregates these test statistics across scales and coordinates. Unlike OCD, which is tailored for detecting changes in the mean of a multi-dimensional Gaussian data stream, a detector from \citet{chan2017optimal} tests the null hypothesis of data following a multivariate normal distribution against an alternative where one of the coordinates has a mixture distribution. 
Such methods constitute state-of-the-art models used to identify changes in high-dimensional data streams, with a focus on changes to the mean, from online changepoint detection literature in statistics. More information on their implementation is provided in the Appendix.

\subsection{Detection Performance Metrics}
\label{section:detection_metrics}
To evaluate the performance of the different detectors, we consider both AUROC as well as average detection time, i.e. the average number of steps before an anomalous episode is classified as out-of-distribution. 
If the anomaly is not classified as OOD before the end of the episode, we set the detection time to the maximum episode length.

\subsection{Detection Performance Results in novel scenarios}
The results for both performance metrics and for all detectors across all scenarios and environments are given in Tables~\ref{tab:ARTS},~\ref{tab:ARNS}, ~\ref{tab:ARNO}.

\paragraph{AUROC results}
We find that DEXTER outperforms all other detectors in the vast majority of scenarios and noise levels, achieving the best performance in 2 out of 2 ARTS settings, 17 out of 18 ARNO settings, and 11 out of 12 ARNS settings. 
We observe that PEDM detector performs no better than random guessing in ARTS setting (average AUROC of 0.51), as well as marginally better than random guessing in ARNO scenarios (average AUROC of 0.59). The main exception is the 1-step correlated noise setting in ARNO Reacher environment, where PEDM performs with an average of 0.84 AUROC across the three noise levels, and outperforms DEXTER on the Light Noise level. 
The performance of PEDM on ARNS is similar, with an average AUROC score of 0.57 across all noise levels and environments.
Such results confirm our hypothesis about the model's limitations outlined in Section \ref{sec:limitations}. 
CPD detectors perform only marginally better than PEDM across the three types of noise in Cartpole and Acrobot, though their AUROC scores across Reacher environments in ARNO and ARNS settings are below PEDM, with an average of around 0.51. The scores of changepoint detectors in these settings are also quite repetitive, since in the vast majority of cases, they consistently detect an anomaly at one of the earliest steps in the episode. Since both algorithms are designed to identify changes to the \textit{mean} of time series data, this may be especially difficult in high-dimension noisy environments, such as Reacher.

\paragraph{Detection time results}
We find that DEXTER+C outperforms PEDM+C by a large margin in the vast majority of settings, in accordance with the higher AUROC scores.

\subsection{Detection Performance Results in Benchmark Scenarios}
We observe in Table~\ref{tab:ARNS}~that DEXTER (DEXTER+C) generally outperforms PEDM (PEDM+C) for minor anomaly scenarios, while the opposite holds true for severe anomalies. This finding is as expected, as DEXTER requires a larger time-series window to extract features. These results suggest that a combination of both DEXTER (DEXTER+C) and PEDM (PEDM+C) approaches might result in optimal outcomes.

\section{Conclusion and future work}

This paper focused on the problem of generalized OOD detection in reinforcement learning. We started by outlining the terminological discussions that are currently present in the literature, and offered a framework that formalizes the problem and aligns its terminology with the broader field of generalized OOD detection in machine learning. We then outlined potential shortcomings in the architectures of current state-of-the-art detectors, and introduced a new set of testing scenarios for generalized OOD detection in RL.

By introducing noise with different types of autocorrelations, these scenarios focus on anomalies that, to the best of our knowledge, have not been explicitly studied in the literature, but are highly relevant to many real-world scenarios. 
We also proposed a new generalized OOD detection model called DEXTER, which first extracts relevant time series features from observations, and then applies an ensemble of isolation forest algorithms to detect potential anomalies. 
Its extension DEXTER+C further establishes a decision rule by use of CUSUM.

Lastly, we also adopted several existing changepoint detection methods from statistics for generalized OOD detection. 
After a series of experiments, we found that DEXTER outperforms the existing state-of-the-art models for OOD detection on the novel test scenarios. 
However, in some benchmark scenarios, a combination of previously used approaches together with DEXTER appears to yield improved results.
In future work, we plan to explore such combinations of different detection mechanisms.

Despite the successes of our method, it still faces several important limitations, which point in the directions to take in future work. First, our evaluations are limited to simulated environments. Future work should test on sim-to-real settings. Second, we do not test on scenarios with noise that is correlated across dimensions. Future work should address cross-dimensional feature extraction. Third, DEXTER uses a fixed window length to detect anomalies, which is fixed before interacting with the test-time environment. Future work should study how this can be replaced by a sliding scale and/or hierarchical pyramid of window sizes. Lastly, drawing on existing work \citep{teh2021expect}, we foresee the process of effectively selecting relevant features at test-time as a natural step to improve DEXTER's efficiency in high-dimensional environments.

Generalized OOD detection is critical to ensure the safety of RL algorithms when they are deployed in the real world. Unlike the kinds of out-of-distribution perturbations that have been considered in previous work, the robotic systems such as drones that have already been used in the real world~\cite{kaufmann.etal2023} risk facing noise that is correlated across timepoints. This noise could come from physical disturbances such as a malfunctioning robot joint or a partially broken camera lens. If the disturbance at each individual time point is slight, it risks not to be caught by systems relying on one-step transitions like PEDM. 

However, over a long time period, these disturbances can cause significant damage to the system, decreasing performance and putting potentially dangerous strain on the rest of the robot. Further, not addressing temporally correlated noise risks leaving the control system open to adversarial attacks, which induce perturbations that are correlated across time, such 
as illusory attacks~\cite{franzmeyer2022illusionary}. This suggests that inclusion of a DEXTER-like system for detection of temporally-correlated noise -- possibly in conjunction with a PEDM-like system optimized at detecting one-step perturbations -- is necessary to ensure safe deployment of RL systems.

More broadly, we hope that our work helps shift future OOD dynamics detection towards more general approaches, improving both AI safety and security. Importantly, this work helps facilitate the move from individual time-step decision rules to information-theoretic optimal CUSUM detection methods.




\begin{acks}
We'd like to thank James Duffy for an in-depth overview of out-of-distribution detection methods in econometrics and statistics. We further thank Philip Torr for helpful comments. LN was supported by the Center for AI Safety (CAIS) and Future of Life Institute for this project. KS was supported by a Cusanuswerk Doctoral Fellowship. CSDW received generous support from the Cooperative AI Foundation. This project received funding by Armasuisse Science+Technology.
\end{acks}


\bibliographystyle{ACM-Reference-Format} 
\bibliography{sample}

\appendix
\input{appendix}


\end{document}

%% file: appendix.tex
\newpage
\section{Appendix}
\subsection{Novel environment definition}
Algorithms 3, 4, and 5 display pseudocodes for generating ARTS, ARNO, and ARNS environments, respectively. 

\begin{algorithm}[h]
\caption{Pseudocode for ARTS environment}

\begin{algorithmic}[1]
\label{pseudocode:ARTS}
\Statex \underline{\textbf{Initialization at \( t = 0 \):}} \\
generate \Call{AR\_noise}{} of size \((num\_dimensions = 1, max\_steps)\)

\Statex \underline{\textbf{Agent step}}
\State sample action \( a_t \sim \pi_{A}(a_t | s_t) \)

\Statex \underline{\textbf{Environment step}}

\Function{Step}{$a_t$}
    \State \( o_{t+1} \) \( \leftarrow \) \Call{AR\_noise}{t}
    \State \Return \( o_{t+1} \), \( r_t \), \( terminated \)
\EndFunction
\end{algorithmic}
\end{algorithm}

\begin{algorithm}[h]
\caption{Pseudocode for ARNO environment}
\label{pseudocode:ARNO}
\begin{algorithmic}[1]
\Statex \underline{\textbf{Initialization at \( t = 0 \):}} \\
generate \Call{AR\_noise}{} of size \((num\_dimensions, max\_steps)\)

\Statex \underline{\textbf{Agent step}}
\State sample action \( a_t \sim \pi_{A}(a_t | s_t) \)

\Statex \underline{\textbf{Environment step}}
\Function{step}{$a_t$}
    \State $s_{t+1}$, $r_t$, $terminated$ $\leftarrow$ \Call{environment.step}{$a_t$, $s_t$}
    \State $o_{t+1}$ $\leftarrow$ $s_{t+1}$ + \Call{AR\_noise}{t}
    \State \Return $o_{t+1}$, $r_t$, $terminated$
\EndFunction
\end{algorithmic}
\end{algorithm}

\begin{algorithm}[h]
\caption{Pseudocode for ARNS environment}
\label{pseudocode:ARNS}
\begin{algorithmic}[1]
\Statex \underline{\textbf{Initialization at \( t = 0 \):}} \\
generate \Call{AR\_noise}{} of size \((num\_dimensions, max\_steps)\)

\Statex \underline{\textbf{Agent step}}
\State sample action \( a_t \sim \pi_{A}(a_t | s_t) \)
\Statex \underline{\textbf{Environment step}}

\Function{step}{$a_t$}
    \State $s'_t$ $\leftarrow$ $s_t$ + \Call{AR\_noise}{t}
    \State $s_{t+1}$, $r_t$, $terminated$ $\leftarrow$ \Call{environment.step}{$a_t$, $s'_t$}
    \State \Return $s_{t+1}$, $r_t$, $terminated$
\EndFunction
\end{algorithmic}
\end{algorithm}

\subsection{Out-of-Distribution Detection Terminology}\label{app:ood-notation}
\citet{muller2022towards} refer to \textit{anomaly detection} as the general task of identifying datapoints that substantially differ from normality in RL environments. The authors do not use the term out-of-distribution (OOD) detection at all.  \citet{danesh2021out} coin the term \textit{out-of-distribution dynamics (OODD) detection} to define the general task of "\textit{detecting when the dynamics of a temporal process change compared to the training-distribution dynamics}". 
The authors differentiate between \textit{sensor-injected} anomalies, which "corrupt the environment observations received by the agent", and \textit{dynamics-injected} anomalies, which \textit{"directly change the dynamics of the environment by modifying key physical parameters of the environment simulator"}. 

\citet{haider2023out} provide a more comprehensive review of terminology issues in the field of OOD detection in reinforcement learning. 
To our knowledge, they also propose the most recent re-definition of out-of-distribution in reinforcement learning as \textit{"severe perturbations of the Markov Decision Process, which effectively change the semantics of the system"}, instead of simply changing the observations that the agent receives. 
The authors draw on the aforementioned review by \citet{yang2021generalized} to argue that a semantic anomaly should effectively shift the transition function of an MDP, i.e. introduce new semantic concepts or changing the environment dynamics. 
In effect, this corresponds to renaming \textit{dynamics-injected} anomalies from \citet{danesh2021out} to \textit{semantic anomalies}. 

While we try to follow the existing terminologies, we believe that the most recent definitions of OOD detection offered by \citet{danesh2021out} and \citet{haider2023out} could be improved for several reasons.
First, \citet{haider2023out} define OOD detection as the identification of semantic anomalies, but do not provide a meaningful term for detecting anomalies that target the observations that the agent receives, which has been a focus of many previous works in the literature on OOD detection in RL \citep{sedlmeier2019uncertainty} \citep{mohammed2021benchmark} \citep{danesh2021out}.

Second, in many real-world environments, reinforcement learning agents could face both sensory and semantic anomalies simultaneously. 
We can take the example of a self-driving car that is trained to drive in normal conditions, and suddenly exposed to a heavy hailstorm. 
On the one hand, the ice will change the observations that the car receives from its cameras, introducing observational noise. 
On the other, the ice will also make the road more slippery, effectively changing the environment dynamics. 
However, the terminology offered by \citep{haider2023out} does not allow to effectively distinguish between these anomalies. 
We think it is reasonable to suggest there should be a common terminology to differentiate between these two types of anomalies in the environment, and that ideally, a well-performing detector should be able to detect both of them. 

Lastly, while the terminology of \citet{danesh2021out} offers a distinction between environmental anomalies, the suggested terms do not correspond to the terminology used by OOD detection outside of reinforcement learning. 
As we discuss in Section \ref{sec:terminology}, it is relatively simple to align the labels for anomalies in reinforcement learning with the standardized terminology from \citet{yang2021generalized}. 

For this reason, Section \ref{sec:terminology} proposes a clarification of terminology for OOD detection in reinforcement learning, which aligns it with literature from other machine learning domains. 

\subsection{Implementation details}
\subsubsection{ARTS, ARNO, ARNS scenarios}
To generate noise with varied orders of correlation, we use the implementation of Autoregressive Process from \texttt{statsmodels}\footnote{More information about the library can be found here: https://www.statsmodels.org/stable/index.html} library in Python. 

\subsubsection{ARNS Acrobot environment}
We implement ARNS scenarios with Light, Medium, and Strong levels of noise on Cartpole and Reacher, as the implementation on Acrobot leads to inconsistent agent policies. In the implementation of Acrobot environment used in this project \citep{gym}, adding additional noise to the underlying states tends to increase the agent reward, which is inconsistent with our definition of Light, Medium, and Strong noise in terms of the reduction in the average cumulative average reward over an episode. 

\subsubsection{Changepoint detectors}
To implement the changepoint detectors from \citet{chen2022high} and \citet{chan2017optimal}, we use the \texttt{ocd} package\footnote{More information about the \texttt{ocd} package can be found at: \href{https://cran.r-project.org/web/packages/ocd/index.html}{https://cran.r-project.org/web/packages/ocd/index.html}} for R, published by \citet{chen2022high}. The package contains the implementation for both detectors. Both algorithms rely on the patience parameter, which is defined as the average run length under the null hypothesis \citep{chen2022high}. Following the definition, we set patience to the average length of the episode under uncorrelated noise in a given setting. For detection threshold, we use the option of automatically calculating the threshold using Monte Carlo simulations, implemented in the \texttt{ocd} package.

Since this project is built on Python codebase, we also had to adapt these detectors from R to Python. Therefore, a small contribution that we provide is, to our knowledge, the first implementation of the algorithms of \citet{chen2022high} and \citet{chan2017optimal} in Python.

\subsection{AUROC results} When reporting the performance of detectors using the Area under the Receiver Operator Characteristic (AUROC), we consider a two-sided test. 
In other words, the victim agent tests whether the predicted anomaly scores are either higher or lower than the anomaly scores predicted for the unperturbed observations. 
Therefore, we report the AUC score as $max(AUROC, 1-AUROC)$.